\documentclass[10pt,twocolumn,letterpaper]{article}

\usepackage[pagenumbers]{cvpr}

\usepackage{xcolor}
\usepackage{url}
\usepackage{booktabs}
\usepackage{multirow}
\usepackage{graphicx}
\usepackage{wrapfig}
\usepackage{tikz}
\usepackage{algorithm}
\usepackage{algpseudocode}
\usepackage{makecell}
\usepackage{dsfont}
\usepackage{colortbl}
\usepackage{changepage}

\usepackage{amsmath,amsfonts,bm}

\def\eqref#1{equation~\ref{#1}}

\def\1{\bm{1}}

\DeclareMathAlphabet{\mathsfit}{\encodingdefault}{\sfdefault}{m}{sl}
\SetMathAlphabet{\mathsfit}{bold}{\encodingdefault}{\sfdefault}{bx}{n}

\DeclareMathOperator*{\argmin}{arg\,min}

\usetikzlibrary{positioning,arrows.meta,shapes,calc,backgrounds}
\usepackage{siunitx}

\definecolor{cvprblue}{rgb}{0.21,0.49,0.74}
\usepackage[pagebackref,breaklinks,colorlinks,allcolors=cvprblue]{hyperref}

\def\paperID{8134}
\def\confName{CVPR}
\def\confYear{2025}

\title{Meta-Learning Hyperparameters for Parameter Efficient Fine-Tuning}

\author{%
    Zichen Tian$^1$ \quad Yaoyao Liu$^2$ \quad Qianru Sun$^1$ \\[8pt]
    $^1$Singapore Management University\\
    $^2$University of Illinois Urbana-Champaign\\[8pt]
    {\small \texttt{zichen.tian.2023@phdcs.smu.edu.sg, lyy@illinois.edu, qianru.sun@smu.edu.sg}}
}

\begin{document}
\maketitle
\begin{abstract}
    \label{sec_abstract}
    Training large foundation models from scratch for domain-specific applications is almost impossible due to data limits and long-tailed distributions — taking remote sensing (RS) as an example.
    Fine-tuning natural image pre-trained models on RS images is a straightforward solution. To reduce computational costs and improve performance on tail classes, existing methods apply parameter-efficient fine-tuning (PEFT) techniques, such as LoRA and AdaptFormer. However, we observe that fixed hyperparameters — such as intra-layer positions, layer depth, and scaling factors, can considerably hinder PEFT performance, as fine-tuning on RS images proves highly sensitive to these settings.
    To address this, we propose \texttt{MetaPEFT}, a method incorporating adaptive scalers that dynamically adjust module influence during fine-tuning. \texttt{MetaPEFT} dynamically adjusts three key factors of PEFT on RS images: module insertion, layer selection, and module-wise learning rates, which collectively control the influence of PEFT modules across the network. We conduct extensive experiments on three transfer-learning scenarios and five datasets in both RS and natural image domains. The results show that \texttt{MetaPEFT} achieves state-of-the-art performance in cross-spectral adaptation, requiring only a small amount of trainable parameters and improving tail-class accuracy significantly. \emph{Our code is available at \url{https://github.com/doem97/metalora}.}

\end{abstract}
\section{Introduction}
\label{sec_intro}
Training large foundation models from scratch for domain-specific applications presents fundamental challenges, particularly in domains with limited data availability and severe data imbalance issues~\cite{wang3d}. 
Take the remote sensing (RS) domain as an example.
RS image recognition has critical applications in environmental monitoring, resource management, and disaster response~\cite{zhu2017deep,ma2019deep,zhang2024vision}. However, learning large foundation models (\textit{e.g.}, CLIP~\cite{clip} and Stable Diffusion~\cite{rombach2022high}) on RS data presents three fundamental challenges. 
First, RS data exhibits high spectral diversity, encompassing different spectral bands from optical remote sensing (ORS, $400$-$700$nm) to synthetic aperture radar (SAR, $1$mm-$1$m), which capture distinct physical properties of Earth's surface~\cite{shaw2003spectral,sarhandbook}. This spectral diversity poses challenges for developing generic, cross-band models~\cite{wang2022self,hong2024spectralgpt}. 
Second, although large foundation models are prevalent in natural image domains, it is impractical to train dedicated foundation models for each RS spectral band due to data scarcity~\cite{sarhandbook,zhu2021deep,guo2023skysense} and the prohibitive computational resources required for training or fine-tuning~\cite{cong2022satmae}. 
Third, RS data often displays a long-tailed distribution~\cite{lam2018xview,xia2018dota,zhang2021shiprsimagenet,christie2018functional}, which causes fine-tuned foundation models to overfit to limited training samples in tail classes, resulting in poor generalization~\cite{tian2024deblora}.

\begin{figure*}[t]
    \centering
    \includegraphics[width=\textwidth]{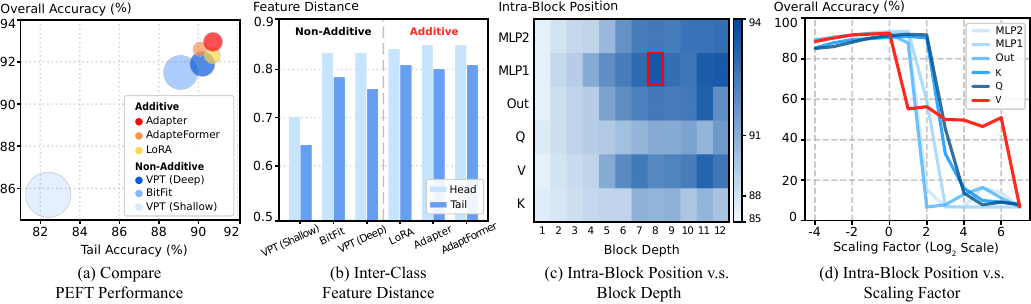}
    \caption{
        \textbf{Comparing PEFT methods for the model adaptation of IN21K $\to$ DOTA.}
        (a) Bubble plot of overall accuracy, tail-class accuracy, and performance variance (in standard deviation) of 5 PEFT methods in 6 total versions. Additive methods exhibit consistently higher accuracy and lower variance than non-additive methods.
        (b) Inter-class feature distances of the PEFT methods measured by cosine similarity. Additive methods achieve 13\% further feature distances (which means better discrimination among tail classes) with comparable head-class distances.
        (c) Accuracy heatmap for applying PEFT on different positions of ViT: on different intra-block layers v.s. among different attention blocks (depth). Deeper blocks yield better performance (86.5\% to 90.4\%), but the combination of optimal block and intra-block position shows unexpected degradation (0.6\% drop for FFN layer with depth 10$\to$11). The \includegraphics[height=.7em]{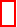} marks optimal combination.
        (d) Accuracy heatmap of intra-block positions v.s. scaling factors. Different positions show distinct sensitivity to scaling factors, with sharp accuracy drops observed (\eg, applying PEFT on the attention output layer (denoted as Out) drops from 87\% to 6.7\% when its PEFT scaling factor increases from 2 to 4). These highlight the non-monotonic complexity of PEFT hyperparameters.
    }
    \label{fig_teaser}
\end{figure*}

Adapting models from natural image domains to RS domains has become a more feasible and practical approach than training models from scratch. Compared to the resource-intensive process of full fine-tuning, Parameter-Efficient Fine-Tuning (PEFT) has emerged as a more efficient adaptation method for foundation models. 
PEFT includes various techniques, such as LoRA~\cite{hu2021lora}, AdaptFormer~\cite{Chen2022adaptformer}, and BitFit~\cite{zaken2022bitfit}. The common idea is to update only a small subset of parameters while keeping the pre-trained weights of the foundation models frozen. 
This approach reduces computational and data requirements, preserves general knowledge learned in natural image domains~\cite{xin2024parameter}, and reduces overfitting in tail classes~\cite{shi2024lift,tian2024deblora}.

In this paper, we conduct a comprehensive study of various PEFT methods in the RS domain and propose our improved method called \texttt{MetaPEFT} that is generic and more efficient than any individual technique of PEFT.
Specifically, we study five representative PEFT methods:\footnote{We provide detailed PEFT taxonomy and our classification criteria of LoRA in Section~\ref{sec_related}.} a selective method called BitFit~\cite{zaken2022bitfit}; three additive methods respectively called LoRA~\cite{hu2021lora}, Adapters~\cite{houlsby2019adapters}, and AdaptFormer~\cite{Chen2022adaptformer}; and a soft-prompt method called VPT~\cite{jia2022vpt}.
We make two key observations.
First, additive methods outperform non-additive methods in both performance and stability.
As shown in Figure~\ref{fig_teaser}(a), additive methods consistently achieve higher overall accuracy and tail-class accuracy while maintaining lower performance variance (represented by bubble size).
Similar observations in the NLP domain~\cite{houlsby2019parameter} and natural image domain~\cite{he2021towards} suggest that such superiority stems from both architectural design (zero-initialization) and hyperparameter choice (scaling factor).
Zero-initialization ensures that parameter updating starts from the pre-trained state, while the scaling factor preserves update directions and modulates only magnitudes.
Second, additive methods achieve better tail-class feature discrimination.
As visualized in Figure~\ref{fig_teaser}(b), additive methods achieve $13$\% higher inter-class feature distances by average (using cosine distance) for tail classes while maintaining comparable distances for head classes.
We attribute this improvement to its flexibility in insertion positions.
For example, the soft-prompt method VPT operates only at input layers. In contrast, additive modules like LoRA and AdaptFormer can be inserted at diverse positions (\eg, across attention blocks of ViT and on the Q/K/V~\cite{hu2021lora} or FFN layers~\cite{Chen2022adaptformer} of any attention block).
Such flexibility enables effective model adaptation without needing large-scale training data,
benefitting tail classes~\cite{kang2019decoupling,yang2020rethinking}.
We thus use additive PEFT as a strong baseline to tackle RS tasks.

We found that additive methods vary mainly based on three key hyperparameters: adaptation position inside an individual attention block, adaptation position across attention blocks (depth), and scaling factors.\footnote{Other relevant but less significant hyperparameters, such as module size and initialization, will be discussed in the Appendix.} For example, LoRA applies its low-rank adaptation to Q/K/V layers, while AdaptFormer applies its adaptation weights to FFN layers. 
We conduct thorough ablation studies on these three hyperparameters and present the resulting heatmap visualizations in Figures~\ref{fig_teaser}(c) and~\ref{fig_teaser}(d).
From the figures, we make two key observations.
1) Model performance exhibits high sensitivity to both hyperparameters.
Specifically, accuracy varies by $86\%$ across scaling factors (from $8.1\%$ to $91.1\%$ on the K projection layer), $4.0\%$ across attention block depths (from $86.5\%$ in block 1 to $90.5\%$ in block 11), and $2.4\%$ across intra-block positions (from $90.4\%$ on K projection layer to $92.8\%$ on FFN layer).
2) Individual hyperparameters demonstrate monotonic trends, \eg, deeper insertion positions generally lead to better performance in Figure~\ref{fig_teaser}(c). Their combinations show complex non-monotonic effects in Figure~\ref{fig_teaser}(d).
In other words, combining individually optimal hyperparameters often leads to unexpected performance degradation,
\eg, using optimal intra-block position (\ie, FFN) with optimal block depth (\ie, depth 11) results in a $0.6\%$ accuracy drop.
Although an exhaustive search might handle these non-monotonic effects, it is computationally infeasible due to the large configuration space of complexity $\mathcal{O}(L|\mathcal{S}|N_\alpha)$ (Section~\ref{sec_manual}).
Moreover, gradient-based optimization is infeasible as PEFT hyperparameter tuning poses a mixed discrete-continuous optimization problem: jointly optimizing discrete positions and continuous scaling factors (elaborated in Section~\ref{sec_manual}).
These challenges motivate us to develop an efficient end-to-end hyperparameter optimization method for PEFT.

Specifically, we propose a meta-learning method called \texttt{MetaPEFT},
by introducing two key designs:
1)~a unified modulator of three PEFT-specific hyperparameters (\ie, intra-attention-block position and attention-block depth on ViT, and the scaling factor of PEFT)
and 2)~a bi-level optimization framework to learn this modulator without needing additional training data.
First, the unified modulator contains a set of scalars, each applied to an individual PEFT position (\textit{e.g.}, a Q/K/V projection layer, attention output layer (denoted as Out), or an FFN layer in the attention block of ViT-B/16), without introducing large overhead (\eg, less than 800 additional parameters for LoRA on ViT-B/16).
This modulator controls both PEFT's insertion position and scaling factor:
when the value of a scalar is close to zero, the PEFT is deactivated at the corresponding position (of the scalar);
when this value is significantly high, its magnitude determines ``how strong PEFT is for the adaptation''.
Introducing this modulator makes gradient-based optimization possible due to its differentiable nature.
Second, we optimize this modulator through a bi-level optimization framework with two alternating loops:
the inner loop trains the model parameters with the modulator (\ie, all scalars) fixed,
and the outer loop optimizes the modulator on a randomly sampled subset of training data in each iteration.
This dynamic sampling exposes optimization to diverse data subsets. It hence prevents overfitting, especially effective for tail classes (\eg, $3.5$ times higher accuracy improvement for tail classes than head classes on the setting of ``SatMAE $\rightarrow$ SAR'', with LoRA as a baseline).
Empirically, we found that direct optimization often leads to numerical instability (\ie, negative values); thus, we apply softplus activation to constrain the modulator to be non-negative.
In summary, our \texttt{MetaPEFT} has three key advantages:
1) it converts the mixed discrete-continuous optimization problem into a differentiable formulation, thereby allowing gradient-descent-based optimization;
2) it automatically discovers the optimal adaptation strength for each position; and
3) its dynamic sampling of meta-learning samples (\textit{i.e.}, validation samples) greatly reduces model overfitting to tail classes.

\section{Related Works}
\label{sec_related}

\noindent
\textbf{Long-tailed Model Adaptation.}
Long-tailed learning approaches in computer vision can be categorized into three main strategies: data manipulation through re-sampling and re-weighting techniques exemplified by cRT~\cite{kang2019decoupling} and BBN~\cite{zhou2020bbn}, representation learning via contrastive learning as demonstrated by PaCo~\cite{cui2021parametric}, and classification objective modification methods such as LDAM~\cite{cao2019learning} and Focal Loss~\cite{ross2017focal}. However, training from scratch often yields sub-optimal results compared to leveraging pre-trained models~\cite{yang2020rethinking}. Recent works explore foundation models like CLIP, where BALLAD~\cite{ma2021simple} and VL-LTR~\cite{tian2022vl} directly fine-tune the entire model, while LPT~\cite{dong2022lpt} incorporates external knowledge. Robust fine-tuning strategies have also been explored to maintain generalization under distribution shifts~\cite{zhu2024robust}. Despite promising results, they face fundamental trade-offs between performance and efficiency~\cite{zhang2023deep,shi2024lift}.

\noindent
\textbf{Parameter-Efficient Fine-tuning (PEFT).}
PEFT enables efficient model adaptation while avoiding the computational overhead of full fine-tuning~\cite{houlsby2019adapters,fu2023effectiveness}. By modifying only a small subset of parameters, PEFT achieves comparable performance with significantly reduced costs~\cite{he2021towards}. Current approaches include additive methods such as adapters~\cite{houlsby2019adapters} and LoRA~\cite{hu2021lora}, and selective methods BifFit~\cite{zaken2022bitfit} or DiffPrune~\cite{guo2020parameter}. However, these methods are highly sensitive to hyperparameter choices, including LoRA's rank, adapter dimensions, and selection criteria~\cite{lawton2023neural,shi2024lift}. Our empirical study reveals that while individual hyperparameters show monotonic trends, their combinations lead to complex, non-monotonic effects. This makes manual optimization both time-consuming and suboptimal, especially given the mixed discrete-continuous nature of these hyperparameters~\cite{chen2023parameter,shi2024lift}. Recent work has further shown that parameter-efficient strategies can effectively enhance group robustness during fine-tuning~\cite{zhu2025project}.

\noindent
\textbf{Meta-Learning for Hyperparameter Optimization.}
Traditional hyperparameter optimization through random search~\cite{bergstra2012random} or Bayesian optimization~\cite{snoek2012practical} struggles with PEFT scenarios, as they ignore structural relationships between network modules and require expensive sequential optimization. While methods like population-based training~\cite{jaderberg2017population} and BOHB~\cite{falkner2018bohb} attempt parallel optimization, they remain computationally intensive. Meta-learning approaches such as MAML~\cite{finn2017model} and Meta-SGD~\cite{li2017meta} demonstrate effective learning-to-learn strategies, with online meta-learning~\cite{finn2019online,liu2023online,liu2024wakening,liu2021adaptive} enabling continuous adaptation. Neural architecture optimization methods, including DARTS~\cite{hanxiao2019darts} and Auto-Meta~\cite{kim2018auto}, successfully apply meta-learning principles. However, existing approaches focus primarily on general hyperparameters~\cite{andrychowicz2016learning} or architecture choices~\cite{zoph2016neural} without addressing PEFT-specific challenges. The complex interdependencies in PEFT methods~\cite{hu2021lora,houlsby2019parameter,he2021towards} create a mixed discrete-continuous optimization problem requiring specialized solutions.

\vspace{-.5em}
\section{Method}
\label{sec_method}
\vspace{-.5em}

\begin{figure*}[t]
    \centering
    \includegraphics[width=\textwidth]{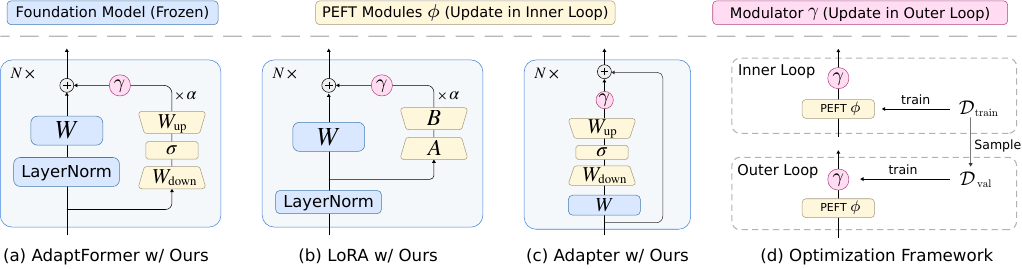}
    \caption{\textbf{Architecture and optimization framework for \texttt{MetaPEFT}.}
        \textbf{In Figures (a)-(c)}, we illustrate how our proposed modulator $\gamma$ is integrated with three representative additive PEFT methods:
        (a) AdaptFormer with modulated up/down projections ($W_\mathrm{up}$/$W_\mathrm{down}$),
        (b) LoRA with modulated low-rank decomposition matrices ($A$/$B$),
        and (c) Adapter with modulated projection layers ($W_\mathrm{up}$/$W_\mathrm{down}$).
        The $\sigma$ denotes the non-linear activation function (\eg, ReLU).
        \textbf{In Figure (d)}, we show our bi-level optimization framework.
        The inner loop optimizes PEFT parameters $\phi$ on training data $\mathcal{D}_\mathrm{train}$, and the outer loop updates modulator $\gamma$ on validation data $\mathcal{D}_\mathrm{val}$ sampled from the training set.
        The symbol $N\times$ indicates the operation is repeated for $N$ attention blocks (\eg, $N$=12 for \texttt{ViT-B/16}).
    }
    \label{fig_model_arch}
\end{figure*}

We have made two key observations in Section~\ref{sec_intro}:
1) additive PEFT methods excel at tail classes, and
2) their hyperparameters exhibit complex non-monotonic combination effects.
In Section~\ref{sec_formulation}, we give a general formulation for additive PEFT methods and elaborate on their specific hyperparameters, such as discrete layer positions and continuous scaling factors. The results of PEFT in Figure~\ref{fig_teaser}(c)-(d) show high sensitivity to the values of hyperparameters. We thus explore hyperparameter optimization.
In Section~\ref{sec_manual}, we
formalize the manual hyperparameter optimization as a mixed integer non-linear programming (MINLP) problem, which is impractical to solve for deep learning models.
In Section~\ref{sec_meta}, we propose our approach, \texttt{MetaPEFT}. We unify both discrete and continuous hyperparameters by converting them into scalars, each applied independently to a specific network position, and optimize them end-to-endly.

\subsection{PEFT and Its Hyperparameters}
\label{sec_formulation}

\noindent
\textbf{Parameter Efficient Fine-Tuning (PEFT).}
Given a pre-trained model parameterized by $\theta$, PEFT transfers pre-trained representations by updating only a small set of parameters $\phi$, \ie, its dimension is usually $\mathrm{dim}(\phi) \ll \mathrm{dim}(\theta)$:
\begin{equation}
    \label{eq_general_peft}
    y=f(x;\theta,\phi),
\end{equation}
where $x$ is the input and $y$ is the output representation. $\theta$ is kept frozen during updating, and only $\phi$ is updated.
In this work, we focus on additive PEFT methods for their high effectiveness (clarified in Section~\ref{sec_intro}).
While non-additive methods modify fixed parameters in the pre-trained model (\ie, $\phi\in\theta$),
additive methods introduce new parameters $\phi$ to carefully selected positions in the model:
\begin{equation}
    \label{eq_additive_peft}
    y=f(x;\theta) + \mathds{1}_p (\alpha \cdot \Delta(x;\phi)),
\end{equation}
where $x \in \mathbb{R}^d$ is the intermediate feature at position $p$,
$f(\cdot)$ is the original layer operation (\eg, linear transformation, attention projections or MLP) with parameters $\theta$,
$\Delta(x;\phi)$ is the additive PEFT module parameterized by $\phi$,
$\alpha$ is the scaling factor,
and $\mathds{1}_p \in \{0,1\}$ indicates whether the PEFT module is used (1) or not (0) at position $p$.
This formulation reveals why additive PEFT achieves improved overall performance and tail-class feature discrimination (Section~\ref{sec_intro}):
$\mathds{1}_p$ allows adaptation at crucial positions, while $\alpha$ precisely controls the adaptation strength, enabling effective feature adjustment, especially for data-scarce tail classes.

Formula~\ref{eq_additive_peft} is for general additive PEFT methods. Different methods have distinct $\Delta(x;\phi)$ designs, as shown in Figure~\ref{fig_model_arch} (a)-(c) in yellow color. For example, LoRA~\cite{hu2021lora} introduces low-rank decomposition of weight $W$ (\ie, $\phi = \{A, B\}$), which can be formulated as:
\begin{equation}
    \label{eq_lora}
    y = (x \cdot W) + \mathds{1}_p (\alpha \cdot x \cdot BA^\top),
\end{equation}
and Adapter/AdaptFormer introduce up/down projections (\ie, $\phi=\{W_\mathrm{down}, W_\mathrm{up}\}$) and activation $\sigma$, formulated as:
\begin{equation}
    \label{eq_adapter}
    y= f(x; \theta) + \mathds{1}_p \left( \alpha \cdot W_{\mathrm{up}} \cdot \sigma\left(W_{\mathrm{down}} \cdot x\right) \right).
\end{equation}

\noindent
\textbf{Block Depth $d$ and Intra-block Position $s$.}
In transformer architectures, the position $p$ in Formula~\ref{eq_additive_peft} consists of two discrete dimensions:
block depth $d\in\{1,...,L\}$ and intra-block position $s\in S$,
where $L$ is the total number of layers (\eg, $12$ for ViT-B/16)
and $\mathcal{S}$ is the set of insertion positions within a transformer block (\ie, Q/K/V projection layers, attention output layer, and FFN layer). Therefore, $\mathds{1}_p$ can be also written as $\mathds{1}_{d,s}$.
Deeper $d$ and $s$ generally yield higher accuracy with improved tail-class separation (Figure~\ref{fig_teaser}(c)).
However, these discrete parameters cannot be optimized through gradient descent,
yet manual hyperparameter optimization often leads to unexpected performance drops (Figure~\ref{fig_teaser}(c)).
In the following subsections, we will introduce an equivalent way of optimizing them but in an end-to-end manner.

\noindent
\textbf{Scaling Factor $\alpha$.}
The scaling factor $\alpha$ in Formula~\ref{eq_additive_peft} is a continuous hyperparameter with a range of $(0,\infty)$.
Empirically, $\alpha$ is usually set within $[0.001,1]$ or start searching from $0.1$~\cite{Chen2022adaptformer}.
It is a hyperparameter used in the forward process of the model training and controls the weight of adding the PEFT representations to the original pre-trained features. In the Appendix, we justify its relationship with the learning rate.
Please note that we do not meta-learn the learning rate as they are not specific to PEFT.

\subsection{Limitations of Manual Optimization}
\label{sec_manual}

Our empirical study (Section~\ref{sec_intro}) has shown that the joint optimization of hyperparameters exhibits complex, non-monotonic interactions.
This makes heuristic or exhaustive search inefficient and challenging.
In this section, we further theoretically justify that this optimization objective is intractable,
and identify its core challenge as a mixed discrete-continuous (MINLP) problem.

\noindent\textbf{Optimization Objective.}
Manual optimization performs exhaustive evaluations of configurations.
Its optimization objective can be formulated as follows:
\begin{equation}
    \label{eq_optim}
    \begin{aligned}
        \max_{d,s,\alpha} \quad & \mathcal{A}_\mathrm{val}(\phi^*_{d,s,\alpha}; \mathcal{D}_\mathrm{val}),                                       \\
        \mathrm{s.t.} \quad     & d \in \{1,...,L\}, s \in \mathcal{S}, \alpha > 0,                                                              \\
                                & \phi^*_{d,s,\alpha} = \argmin_{\phi} \mathcal{L}_\mathrm{train}(\phi, d,s,\alpha; \mathcal{D}_\mathrm{train}),
    \end{aligned}
\end{equation}
where $\phi^*_{d,s,\alpha}$ denotes the optimal PEFT parameters under the given hyperparameter configuration $\{d,s,\alpha\}$. $\mathcal{A}_{\mathrm{val}}$ denotes the validation accuracy, and $\mathcal{L}_{\mathrm{train}}$ denotes the training loss (Logit Adjustment~\cite{menon2020long} loss, in our case).

\noindent\textbf{MINLP Problem.}
The optimization objective Formula~\ref{eq_optim} forms a Mixed Integer Non-Linear Programming (MINLP) problem.
This poses two key challenges.
First, the mix of continuous variable $\alpha$ and discrete variables $d, s$ makes this problem non-differentiable, preventing the direct use of gradient-descent-based optimization methods.
We also prove this problem is NP-hard in the Appendix.
Second, the problem has a large discrete configuration space, \ie, $\mathcal{O}(L|\mathcal{S}|N_\alpha)$.
This makes exhaustive search approaches computationally infeasible,
\eg, manually optimizing ViT-B/16 ($L$=$12$, $|\mathcal{S}|$=$5$) with only $N_{\alpha}$=$10$ requires exploring $600$ configurations.

\subsection{Auto Optimization via Meta-Learning}
\label{sec_meta}
To address this mixed discrete-continuous challenge, we propose to unify all hyperparameters into a differentiable modulator that can be optimized through gradient.

\paragraph{Unified Modulator.}
We unify the positional indicator $\mathds{1}_p$ and scaling factor $\alpha$ (in Equation~\ref{eq_additive_peft}) into a single differentiable modulator $\gamma \in \mathbb{R}$, as illustrated by pink in Figure~\ref{fig_model_arch}:
\begin{equation}
    \label{eq_unified_peft}
    y = f(x;\theta) + \gamma \cdot \Delta(x;\phi).
\end{equation}
This design turns the MINLP problem (Equation~\ref{eq_optim}) into a continuous optimization.
Concretely, when $\gamma \approx 0$, it suppresses the PEFT module's contribution to zero, functionally equivalent to $\mathds{1}_p = 0$;
when $\gamma > 0$, it controls both module activation (\ie, continuous relaxation of $\mathds{1}_p$) and update magnitude (role of $\alpha$).
In practice, we initialize $\gamma=1.0$ to preserve the model's pre-trained behavior during the first inner loop (note that while PEFT modules' zero-initialization preserves pre-trained behavior at the first training step, $\gamma$ remains frozen during the first inner loop),
and adopt softplus activation to ensure non-negativity and numerical stability of $\gamma$.

\paragraph{Bi-Level Optimization Framework.}
The continuous $\gamma$ allows optimization through gradients on the validation set.
We reformulate the optimization objective~\ref{eq_optim} as follows:
\begin{equation}
    \label{eq_meta_optim}
    \begin{aligned}
        \min_{\gamma \in \mathbb{R}^+} \quad & \mathcal{L}_\mathrm{LA}(\phi^*_{\gamma}; \mathcal{D}_\mathrm{val}),                                 \\
        \mathrm{s.t.} \quad                  & \phi^*_{\gamma} = \argmin_{\phi} \mathcal{L}_\mathrm{LA}(\phi, \gamma; \mathcal{D}_\mathrm{train}),
    \end{aligned}
\end{equation}
where $\mathcal{L}_\mathrm{LA}$ denotes the Logit Adjustment~\cite{menon2020long} loss.
We solve this bi-level optimization by alternating inner loop
\begin{equation}
    \label{eq_inner_update}
    \phi_{t+1} = \phi_t - \eta_{\phi} \nabla_{\phi} \mathcal{L}_\mathrm{LA}(\phi_t, \gamma_t; \mathcal{D}_\mathrm{train})
\end{equation}
and outer loop
\begin{equation}
    \label{eq_outer_update}
    \gamma_{t+1} = \gamma_t - \eta_{\gamma} \nabla_{\gamma} \mathcal{L}_\mathrm{LA}(\phi_{t+1}; \mathcal{D}_\mathrm{val}),
\end{equation}
where $\eta_{\phi}$ and $\eta_{\gamma}$ are learning rates for inner and outer loops, respectively.
The inner loop optimizes PEFT parameters with frozen modulator $\gamma$, while the outer loop updates modulator $\gamma$ with frozen PEFT parameters.
In practice, for each outer loop, we randomly sample $20\%$ of training data as $\mathcal{D}_\mathrm{val}$.
This sampling strategy serves as an implicit regularization, \ie, it reduces hyperparameter overfitting by exposing the optimization to diverse data subsets.
This strategy particularly benefits tail classes where overfitting is severe.
Besides, we alternate between inner/outer loops every $K$ inner steps ($K$ is discussed in Section~\ref{sec_experiments}).
This alternating strategy balances sufficient optimization of PEFT parameters while ensuring timely modulator updates.
\vspace{-.5em}
\section{Experiments}
\label{sec_experiments}
\vspace{-.2em}

\subsection{Model Adaptation Scenarios}
\label{sec_settings}

To comprehensively evaluate our method's effectiveness in addressing core challenges in RS and natural image domains, we design experiments in three transfer-learning scenarios. These scenarios systematically validate our method's capability in handling spectral diversity, data scarcity, and long-tailed distribution issues in both RS and natural image domains.

\noindent\textbf{Natural Vision Domain Adaptation.}
In the natural image domain, we use three standard long-tailed benchmarks: CIFAR100-IR100~\cite{krizhevsky2009learning} ($50$K samples, $100$ classes, imbalance ratio $100$), Places-LT~\cite{liu2019large,zhou2017places} ($62$K samples, $365$ classes, imbalance ratio $996$), and iNaturalist-2018~\cite{van2018inaturalist} ($437.5$K samples, $8,142$ species, imbalance ratio $500$).
Each dataset has varying imbalance ratios, thus can help establish our method's baseline performance on conventional long-tailed benchmarks.

\noindent\textbf{Natural Vision to RS Domain Adaptation.}
To evaluate our method's cross-domain adaptation and data scarcity handling abilities, we transfer an IN21K pre-trained ViT-B/16 model to one representative optical remote sensing (ORS) dataset, DOTA~\cite{dota1,dota2,dota3}, which contains 188K images across $15$ object categories.
DOTA exhibits a significant imbalance ratio of 86, with class distribution (in the whole dataset) ranging from $28.35\%$ (ship) to $0.64\%$ (helicopter).
Following standard practice, we categorize the classes into the head ($6$ classes, $>$$5\%$ samples), middle ($3$ classes, $1$-$5\%$), and tail ($6$ classes, $<$$1\%$) groups.

        \noindent\textbf{Cross-Spectral RS Domain Adaptation.}
        To validate our method's effectiveness in handling spectral diversity, we transfer the optical pre-trained model, SatMAE-L/16 (non-temporal edition)~\cite{cong2022satmae}, to the synthetic aperture radar (SAR) domain using the FUSRS v2 dataset~\cite{tian2024deblora}.
        This dataset combines FUSAR-Ship~\cite{hou2020fusar} and SRSDD~\cite{lei2021srsdd} datasets, containing 5,261 high-resolution ($1$m-$10$m) SAR ship images for training.
        We use categories with $<20\%$ training samples as tail classes.
        The large domain gap between ORS and SAR, coupled with the small dataset and class imbalance, makes this adaptation scenario particularly challenging.

        \subsection{Implementation Details}
        \label{sec_implementation}

        Our \texttt{MetaPEFT} alternates between training PEFT modules and learning their modulators through a bi-level optimization framework.
        Here, we detail the implementation of both alternate steps.

        \noindent\textbf{PEFT Configuration.}
        We adopt a principled approach~\cite{clip} to determine PEFT module sizes based on dataset complexity.
        We control the size of the PEFT module based on setting up different ranks in LoRA or different adapter dimensions in Adapter/AdaptFormer. For simplicity, we use the term ``PEFT dimension'' in the following texts to represent the rank/dimension.
        For natural vision datasets, we set PEFT dimension proportional to the number of classes while maintaining parameter efficiency: rank $4$ for CIFAR100-IR100 ($100$ classes), rank $8$ for Places-LT ($365$ classes), and rank $256$ for iNaturalist-2018 ($8,142$ classes).
        For RS datasets, considering their domain-specific characteristics and limited data, we empirically set the PEFT dimension as $4$ after validation experiments showing little gains from using larger PEFT dimensions.

        \begin{table*}[h]
    \centering
    \small
    \begin{tabular}{lccccccccccccccc}
        \toprule[1.3pt]
        \multirow{2}{*}{Position} & \multicolumn{3}{c}{CIFAR100} & \multicolumn{3}{c}{iNat2018} & \multicolumn{3}{c}{Places-LT} & \multicolumn{3}{c}{DOTA} & \multirow{2}{*}{Avg} & \multirow{2}{*}{Avg$_\mathrm{tail}$}                                                                                                                                  \\
        \cmidrule(lr){2-4} \cmidrule(lr){5-7} \cmidrule(lr){8-10} \cmidrule(lr){11-13}
                                  & Head                         & Med                          & Tail                          & Head                     & Med                  & Tail                                 & Head          & Med           & Tail          & Head          & Med           & Tail           &               &               \\
        \midrule \midrule
        ATTN                      & 92.2                         & 87.7                         & 86.3                          & \textbf{67.9}            & 76.3                 & 76.7                                 & 48.4          & 48.3          & \textbf{45.7} & 94.1          & 94.7          & 91.1           & 75.8          & 74.9          \\
        FFN                       & \textbf{92.4}                & 87.9                         & \textbf{86.8}                 & 66.7                     & 75.6                 & 76.9                                 & \textbf{48.6} & 47.9          & 45.6          & \textbf{94.6} & 94.8          & 91.4           & 75.8          & 75.2          \\
        ATTN-FFN                  & 92.3                         & \textbf{88.1}                & 86.6                          & 67.7                     & \textbf{76.8}        & \textbf{77.5}                        & 48.2          & \textbf{48.3} & 45.5          & \textbf{94.6} & \textbf{95.4} & \textbf{92.4 } & \textbf{76.1} & \textbf{75.5} \\
        \bottomrule[1.3pt]
    \end{tabular}
    \caption{
        \textbf{Applying PEFT on Different Intra-Block Layer Positions.}
        We compare three LoRA insertion positions within the attention block: attention-projection-layers-only (ATTN), feedforward-network-only (FFN), and combined (ATTN-FFN) across four adaptation settings: IN21K pre-trained ViT-B/16 $\to$ \{CIFAR100, iNaturalist2018, Places-LT, DOTA\}.
        Results reported in Accuracy ($\%$).
        Results show that combined positions yield slightly better overall performance and tail-class accuracy.
    }
    \label{tab_module_impact}
    \vspace{-1em}
\end{table*}
        \begin{table}[t]
    \centering
    \small
    \begin{tabular}{llcccc}
        \toprule[1.3pt]
        \multirow{2}{*}{Module} & \multirow{2}{*}{Position} & \multicolumn{4}{c}{Accuracy (\%)}                                                 \\
        \cmidrule(lr){3-6}
                                &                           & Head                              & Med           & Tail          & Avg           \\
        \midrule \midrule
        \multirow{4}{*}{ATTN}   & K                         & 91.6                              & 93.0          & 87.7          & 90.6          \\
                                & Q                         & 91.4                              & 93.1          & 90.0          & 91.5          \\
                                & V                         & 94.0                              & 94.5          & 90.2          & 92.7          \\
                                & Out                       & 93.6                              & 94.1          & 90.0          & 92.3          \\
        \midrule
        \multirow{2}{*}{FFN}    & MLP 1                     & \textbf{94.6}                     & \textbf{94.6} & \textbf{91.6} & \textbf{93.4} \\
                                & MLP 2                     & 93.6                              & 94.3          & 90.6          & 92.7          \\
        \bottomrule[1.3pt]
    \end{tabular}
    \caption{\textbf{Applying PEFT on Layer-Wise Intra-Block Positions for IN21K$\to$DOTA.}
        We compare LoRA's different layer-wise insertion positions within the attention block.
        ATTN in Table~\ref{tab_module_impact} now has $4$ versions due to $4$ kinds of projection layers (Q/K/V/Out) in the attention computation and output of the attention block.
        Results reported in Accuracy ($\%$).
        %
        %
        LoRA on the first FFN layer (MLP 1) yields the best performance with an average accuracy of $93.4\%$.
    }
    \label{tab_position_impact}
    \vspace{-1em}
\end{table}

        \noindent\textbf{Meta Training Protocol.}
        Our training follows the bi-level optimization framework elaborated in Section~\ref{sec_meta}.
        The inner loop optimizes PEFT parameters using SGD with a base learning rate of $\eta_{\phi} = 1\times10^{-2}$ using a Logit Adjustment loss~\cite{menon2020long}.
        We use a base batch size $128$ with square-root scaling~\cite{goyal2017accurate} for learning rate adjustment.
        For the outer (meta) loop, we randomly hold out $20\%$ of training data stratified by class to maintain class distribution.
        The outer loop optimizes our proposed modulator using Adam with the meta-learning rate $\eta_{\gamma}$ (hyperparameter that varies depending on the dataset and backbone) and L2 regularization. It has the same loss function as the inner loop.
        Especially for RS datasets, we make specific adjustments: DOTA uses $20$ inner loop steps with batch size $512$, while FUSRS uses $10$ steps with batch size $256$ and a reduced learning rate due to limited data.
        We implement early stopping when validation accuracy shows less than $0.3\%$ improvement over $3$ consecutive epochs.
        All experiments are conducted on four NVIDIA V100/3090 GPUs, with training times ranging from $2$ hours (CIFAR100-IR100) to $6$ hours (iNaturalist-2018).
        More implementation details are available in the Appendix.

        \noindent\textbf{Fair Comparison.}
        we ensure a fair comparison between low-rank and other additive methods.
        Concretely, we ensure a similar module size between LoRA and adapters by fixing the rank of LoRA and the adapter's hidden dimension to a comparable size.
        We follow the setting in~\cite{shi2024lift} to set higher hidden dimensions for larger classifier heads.
        The learnable parameters in different methods on different datasets are provided in Appendix Table~S2.

        \subsection{Ablation Studies and Method Comparisons}
        \label{sec_ablation}

        Our observations in Section~\ref{sec_intro} identify three key hyperparameters: intra-block position, block depth, and scaling factor.
        In this section, we provide detailed ablation results for each of them and their combinations.

        \noindent\textbf{Impact of intra-block position.}
        We examine the effectiveness of different positions (projection layers) within an attention block.
        As shown in Table~\ref{tab_position_impact}, the MLP1 (\ie, the first FFN layer) achieves the best performance ($93.4\%$ average accuracy).
        It performs particularly better (than others) in tail classes ($91.6\%$).
        The results also show that FFN positions (MLP1, MLP2) outperform attention-related positions (K, Q, V, Out), with performance gaps ranging from $0.7\%$ to $2.8\%$.
        This might be because FFN layers focus on feature transformation, while attention layers primarily handle spatial correlations.
        This feature makes FFN layers more suitable for domain adaptation, where the distribution shift is the main challenge.
        Besides, in Table~\ref{tab_module_impact}, we examine different layer combinations.
        Results show that attention and FFN modules are complementary to each other (\eg, ATTN+FFN obtained $1.3\%$ improvement on DOTA's tail classes).
        We think this complementarity is because attention focuses on global context and FFN processes local features. Combining them results in more robust adaptation.

        \noindent\textbf{Impact of block depth.}
        Figure~\ref{fig_teaser}(c) shows the impact of cumulative last-N blocks. As an extended investigation, in Table~\ref{tab_layer_depth_v}, we show the results of different block groups.
        The middle-lower blocks (L3-5) achieve the best performance ($91.9\%$), followed closely by middle-upper blocks (L6-8).
        Intriguingly, the deepest blocks (L9-11) show the largest performance drop ($3.2\%$).
        This effect is amplified for tail classes, where the gap between optimal and sub-optimal block reaches $4.5\%$.
        These observations suggest PEFT modules should be applied with a stronger scaling factor on middle blocks rather than being applied uniformly or only to deep blocks.
        In addition, we also visualized the effect of using joint hyperparameters in Figure~\ref{fig_teaser}.
        We will provide more detailed results (including head/med/tail performances) in the supplementary materials.

        \noindent\textbf{Random sampling strategy.}
        We study how the proportion of training data in the outer loop impacts performance.
        As shown in Table~\ref{tab_compare_sampling}, increasing the sampling ratio from $10\%$ to $30\%$ leads to consistent performance improvements across all classes, with the most significant gains observed in tail classes (from $90.8\%$ to $93.4\%$).
        This trend suggests that insufficient sampling may lead to sub-optimal modulator optimization.
        We adopt $20\%$ as our default setting in main comparisons, as it provides a good balance between performance ($94.2\%$ average accuracy) and training time.

        \noindent\textbf{Compare with state-of-the-art.}
        We evaluate our method by incorporating it into three representative additive PEFT methods.
        Results in Table~\ref{tab_compare_peft} reveal three key findings.
        \textit{First}, our method shows strong synergy with LoRA, consistently improving its performance across all metrics and achieving the highest average accuracy ($+1.13\%$ to $83.97\%$ for LoRA).
        \textit{Second}, our method demonstrates particular strength in handling large domain gaps.
        In the challenging SatMAE$\to$SAR scenario, where the domain shift is most significant, our method achieves its largest improvement on tail classes ($+1.2\%$ for LoRA).
        This potentially stems from our method's ability to selectively strengthen or weaken different blocks' adaptation based on their domain-specific importance.
        \textit{Third}, the consistent improvements over iNat2018 ($8,142$ classes) demonstrate our method's capability to tackle extreme long-tailed scenarios.
        Even with such extensive class space, our method still provides stable gains ($+0.8\%$ on tail classes).
        This is because our random sampling strategy forces the model to optimize the modulator on diverse subsets of data in each iteration, reducing model overfitting to some extent.

        \begin{table}[t]
    \centering
    \small
    \begin{tabular}{lccccc}
        \toprule[1.3pt]
        Block Group & Many & Med & Few & Avg & Drop (\%) \\
        \midrule \midrule
        L9-11 & 88.1 & 89.8 & 88.8 & 89.0 & 3.2 \\
        L6-8 & 92.3 & 93.5 & 89.3 & 91.6 & 0.3 \\
        L3-5 & \textbf{92.4} & \textbf{93.7} & \textbf{89.8} & \textbf{91.9} & - \\
        L0-2 & 90.5 & 92.8 & 88.4 & 90.6 & 1.4 \\
        \bottomrule[1.3pt]
    \end{tabular}
    \caption{
        \textbf{Ablate Different Block Depths for IN21K$\to$DOTA ($\%$).}
        We analyze the impact of applying LoRA across different transformer blocks.
        %
        Results reported in Accuracy ($\%$).
        Results indicate that using LoRA on the middle-lower blocks (L3-5) achieves the best performance ($91.9\%$ overall accuracy), while the deepest blocks (L9-11) show an unexpected performance drop.
        This suggests that we should apply a higher scaling factor for LoRA in intermediate blocks.
    }
    \label{tab_layer_depth_v}
    \vspace{-1em}
\end{table}

        \begin{table*}[t]
    \centering
    \small
    \begin{tabular}{llcccccccccc}
        \toprule
        \multirow{2}{*}{Category} & \multirow{2}{*}{Method} & \multicolumn{3}{c}{IN21K $\to$ iNat2018} & \multicolumn{3}{c}{IN21K $\to$ DOTA} & \multicolumn{2}{c}{SatMAE $\to$ SAR} & \multirow{2}{*}{Avg$_\text{tail}$} & \multirow{2}{*}{Avg} \\
        \cmidrule(lr){3-5} \cmidrule(lr){6-8} \cmidrule(lr){9-10}
        & & Head & Med & Tail & Head & Med & Tail & Head & Tail & & \\
        \midrule \midrule
        \multirow{3}{*}{Non-additive} & VPT-Shallow & 57.5 & 65.5 & 65.9 & 85.4 & 89.0 & 82.4 & 39.8 & 68.4 & 72.23 & 69.24 \\
        & BitFit & 57.6 & 67.3 & 68.4 & 92.0 & 93.7 & 89.1 & 32.1 & 74.7 & 77.40 & 71.86 \\
        & VPT-Deep & 64.9 & 74.2 & 75.9 & 92.1 & 93.5 & 90.2 & 30.6 & 50.0 & 72.03 & 71.43 \\
        \cmidrule(l){1-12}
        \multirow{6}{*}{Additive} & Adapter & 67.9 & 76.8 & 77.7 & 93.2 & 94.9 & 90.6 & 37.9 & 75.8 & 81.37 & 76.85 \\
        & \cellcolor{gray!20}\quad w/ Ours & \cellcolor{gray!20}68.2 & \cellcolor{gray!20}76.9 & \cellcolor{gray!20}78.1 & \cellcolor{gray!20}93.2 & \cellcolor{gray!20}\textbf{95.1} & \cellcolor{gray!20}90.7 & \cellcolor{gray!20}37.7 & \cellcolor{gray!20}76.0 & \cellcolor{gray!20}81.60 & \cellcolor{gray!20}76.99 \\
        & AdaptFormer & 68.7 & 76.7 & 78.0 & 93.2 & 94.9 & 90.1 & 35.3 & \textbf{76.7} & 81.60 & 76.70 \\
        & \cellcolor{gray!20}\quad w/ Ours & \cellcolor{gray!20}68.7 & \cellcolor{gray!20}76.6 & \cellcolor{gray!20}78.2 & \cellcolor{gray!20}92.7 & \cellcolor{gray!20}94.6 & \cellcolor{gray!20}90.1 & \cellcolor{gray!20}35.5 & \cellcolor{gray!20}76.4 & \cellcolor{gray!20}81.57 & \cellcolor{gray!20}76.60 \\
        & LoRA & 69.1 & 77.3 & 78.5 & 93.1 & 93.4 & 90.7 & 40.0 & 72.1 & 80.43 & 76.78 \\
        & \cellcolor{gray!20}\quad w/ Ours & \cellcolor{gray!20}\textbf{70.2} & \cellcolor{gray!20}\textbf{78.6} & \cellcolor{gray!20}\textbf{79.3} & \cellcolor{gray!20}\textbf{93.9} & \cellcolor{gray!20}\textbf{95.1} & \cellcolor{gray!20}\textbf{91.4} & \cellcolor{gray!20}\textbf{40.6} & \cellcolor{gray!20}74.2 & \cellcolor{gray!20}\textbf{81.63} & \cellcolor{gray!20}\textbf{77.91} \\
        \bottomrule
    \end{tabular}
    \caption{\textbf{Comparing different PEFT methods \emph{w/} or \emph{w/o} ours.}
        We evaluate three transfer scenarios (IN21K$\to$iNat2018, IN21K$\to$DOTA, SatMAE$\to$SAR) using non-additive methods (VPT, BitFit) and additive methods (Adapter, LoRA, AdaptFormer).
        Performance is measured across head, medium, and tail classes for each dataset (the SAR dataset has only head and tail) and reported in Accuracy ($\%$).
        ``Avg$_\text{tail}$'' shows mean performance on tail classes, and ``Avg'' is the macro-average across all class splits.
        Results show that our method consistently improves additive methods, with LoRA achieving the highest tail-class performance ($81.63\%$).
        Ours are marked in \setlength{\fboxsep}{2.5pt}\colorbox{gray!20}{gray}.
    }
    \label{tab_compare_peft}
    \vspace{-.8em}
\end{table*}

        \begin{table}[t]
    \centering
    \small
    \begin{tabular}{llcc}
        \toprule
        \multirow{2}{*}{Category} & \multirow{2}{*}{Method} & \multicolumn{2}{c}{Inter-class Distance} \\
        \cmidrule(lr){3-4}
        & & Head & Tail \\
        \midrule \midrule
        \multirow{3}{*}{Non-additive} & VPT-Shallow & 0.65 & 0.58 \\
        & BitFit & 0.81 & 0.75 \\
        & VPT-Deep & 0.81 & 0.72 \\
        \cmidrule(l){1-4}
        \multirow{6}{*}{Additive} & Adapter & 0.83 & 0.77 \\
        & \cellcolor{gray!20}\quad w/ Ours & \cellcolor{gray!20}0.84 & \cellcolor{gray!20}0.79 \\
        & AdaptFormer & 0.83 & 0.78 \\
        & \cellcolor{gray!20}\quad w/ Ours & \cellcolor{gray!20}0.83 & \cellcolor{gray!20}0.77 \\
        & LoRA & 0.82 & 0.78 \\
        & \cellcolor{gray!20}\quad w/ Ours & \cellcolor{gray!20}\textbf{0.83} & \cellcolor{gray!20}\textbf{0.80} \\
        \bottomrule
    \end{tabular}
    \caption{\textbf{Inter-Class Feature Distance.}
        We measure the inter-class \emph{cosine} distance between head classes, and between tail classes.
        Results are reported on transfer scenario IN21K$\to$DOTA using average cosine distance, where higher values indicate better class separation.
        Our method (marked in \setlength{\fboxsep}{2.5pt}\colorbox{gray!20}{gray}) achieves the best feature separation when combined with LoRA.
    }
    \label{tab_feature_dist}
    \vspace{-1.5em}
\end{table}

        \noindent\textbf{``Are we learning a better representation?''}
        Feature distance analysis in Table~\ref{tab_feature_dist} unveils why our method can improve the tail-class performance.
        First, all additive methods show higher inter-class distances than non-additive ones, the gap is particularly pronounced for tail classes ($0.77$-$0.78$ vs. $0.72$).
        Second, when combined with LoRA, our method achieves the best feature separation ($0.85$ for the head, $0.82$ for the tail) and reduces the performance gap between head and tail class groups (from $0.04$ to $0.03$).
        This balanced improvement stems from our dynamic modulation mechanism, which automatically adjusts adaptation strength across different intra-block positions and block depths.
        Finally, the low distance of VPT-Shallow confirms our assumption that input-only modifications are insufficient for complex domain adaptation.

        \noindent\textbf{Impact of rank in low-rank methods.}
        In Table~\ref{tab_compare_peft}, we found rank $r$ one critical hyperparameter for low-rank methods (\eg, LoRA and its variants).
        It significantly affects the performance and size of PEFT modules.
        Our experiments also demonstrate this rank $r$ is independent of scaler and positions.
        We provide detailed observations about rank selection in the Appendix.

        \begin{table}[t]
    \centering
    \small
    \begin{tabular}{lccccc}
        \toprule[1.3pt]
        Ratio (\%) & Head          & Med           & Tail          & Avg           & $\Delta$Tail (\%) \\
        \midrule \midrule
        5          & 91.2          & 92.1          & 88.2          & 90.5          & -5.2              \\
        10         & 93.4          & 94.2          & 90.8          & 92.8          & -2.6              \\
        30         & \textbf{95.0} & \textbf{95.8} & \textbf{93.4} & \textbf{94.7} & -                 \\
        50         & 94.8          & 95.6          & 93.0          & 94.5          & -0.4              \\
        \bottomrule[1.3pt]
    \end{tabular}
    \caption{
        \textbf{Impact of Sampling Strategies ($\%$).}
        We evaluate different sampling ratios in the outer loop of \texttt{MetaPEFT}.
        Results reported in Accuracy ($\%$).
        While 30\% sampling achieves the best performance (94.7\% average accuracy), we adopt 20\% in our default setting due to time constraints.
    }
    \label{tab_compare_sampling}
    \vspace{-1em}
\end{table}

        \noindent\textbf{Computational overhead.}
        Our \texttt{MetaPEFT} archives $1.13$ percentage points higher accuracy than LoRA (Table~\ref{tab_compare_peft}) with minimal overhead (\textit{e.g.} only $0.0008$M params on ViT-B/16).
The detailed computational overhead is provided in the Appendix.

\vspace{-.5em}
\vspace{-.5em}
\section{Conclusions}
\vspace{-.5em}

In this paper, we identified that additive PEFT methods outperform non-additive ones, but their performance is highly sensitive to insertion positions and scaling factors. We extensively evaluated this phenomenon on five transfer learning scenarios in both RS and natural image domains. To address this challenge, we proposed MetaPEFT, a meta-learning framework that converts discrete and continuous hyperparameters into a unified differentiable modulator optimized through bi-level optimization. Our approach effectively handles the spectral diversity and long-tailed data distributions without requiring extensive manual tuning. We discuss the limitations and future work in the Appendix.

\vspace{-.5em}

\vspace{-.5em}
\section*{Acknowledgments}
\vspace{-.5em}
The authors gratefully acknowledge the support from the DSO research grant awarded by DSO National Laboratories, Singapore. The authors also extend sincere gratitude to Prof. Antoine Ledent (Singapore Management University) for his insightful guidance about the non-convex properties of PEFT optimization during the rebuttal.

    {%
        \small
        \bibliographystyle{ieeenat_fullname}
        \bibliography{main}
    }

\clearpage
\onecolumn
\appendix
\renewcommand{\thesection}{\Alph{section}}
\renewcommand{\theequation}{S\arabic{equation}}
\renewcommand{\thetable}{S\arabic{table}}
\renewcommand{\thefigure}{S\arabic{figure}}

\setcounter{section}{0}
\setcounter{equation}{0}
\setcounter{table}{0}
\setcounter{figure}{0}

\begin{center}
\Large\textbf{Supplementary Material}
\end{center}

\definecolor{darkred}{RGB}{180,0,0}
\definecolor{crimson}{RGB}{220,20,60}

\noindent This supplementary material provides additional theoretical foundations, extended analyses, and experimental results that complement our main manuscript (references to the main manuscript are shown in {\color{darkred}red}):

\textbf{\ref{supp_sec_theoretical_foundations}\@. Theoretical Foundations}
\begin{adjustwidth}{2.5em}{0em}
    \begin{itemize}
        \item Section~\ref{supp_sec_lr_vs_scale} analyzes the relationship between scaling factors and learning rates {\color{darkred}(Section~3.1)}
        \item Section~\ref{supp_sec_np_hardness} proves NP-hardness of PEFT optimization objective {\color{darkred}(Section~3.2, Formulation~6)}
        \item Section~\ref{supp_sec_conv} discusses the non-convex property of our optimization objective {\color{darkred}(Section~3.3)}
    \end{itemize}
\end{adjustwidth}

\textbf{\ref{supp_sec_method_design_analysis}\@. Method Design and Analysis}
\begin{adjustwidth}{2.5em}{0em}
    \begin{itemize}
        \item Section~\ref{supp_sec_outer_loop_sampling} compares various data sampling strategies {\color{darkred}(Section~4.3)}
        \item Section~\ref{supp_sec_more_impl_details} provides implementation specifics, \ie, fair comparison settings, classifier design, and evaluation protocols {\color{darkred}(Section~4.2)}
    \end{itemize}
\end{adjustwidth}

\textbf{\ref{supp_sec_validation_experiments}\@. Validation Experiments and Extended Observations}
\begin{adjustwidth}{2.5em}{0em}
    \begin{itemize}
        \item Section~\ref{supp_sec_lora_pos} generalizes our key observations to natural vision domains {\color{darkred}(Figure~1)}
        \item Section~\ref{supp_sec_position_combinations} analyzes interaction effects between different insertion positions {\color{darkred}(Table~2)}
        \item Section~\ref{supp_grid_search} demonstrates that hyperparameter optimization is particularly crucial for tail classes {\color{darkred}(Figure~1)}
        \item Section~\ref{supp_sec_lora_rank} validates that rank optimization can be decoupled from remaining hyperparameters {\color{darkred}(Section~4.3)}
        \item Section~\ref{supp_sec_tsne} provides t-SNE feature visualization for qualitative comparison {\color{darkred}(Section~4.3)}
        \item Section~\ref{supp_sec_ensemble} evaluates our meta-optimization strategy on more recent low-rank methods. {\color{darkred}(Section~4.3)}
    \end{itemize}
\end{adjustwidth}

\textbf{\ref{supp_sec_limits}\@. Limitations and Future Work}
\begin{adjustwidth}{2.5em}{0em}
    \begin{itemize}
        \item Section~\ref{supp_sec_limits} discusses limitations of our work {\color{darkred}(Section~5)}.
    \end{itemize}
\end{adjustwidth}

\section{Theoretical Foundations}
\label{supp_sec_theoretical_foundations}

\subsection{Equivalence and Advantages of Optimizing Scaling Factors rather than Learning Rate}
\label{supp_sec_lr_vs_scale}
This section motivates our choice of optimizing scaling factors instead of learning rates (supplementary to {\color{darkred}Section~3.1}).
We prove that optimizing scaling factors 1) is equivalent to optimizing the learning rate while 2) offers more flexible position-wise control over PEFT modules.

The fact that ``learning any of them is equivalent'' can be derived from LoRA's mechanism.
We start with LoRA's formulation in the forward pass. The output $y$ of the layer when given input $x$ is:
\begin{equation}
    y = W'x = (W + \alpha(AB))x,
\end{equation}
where $W'$ denotes the updated weight matrix, $W$ is the original pre-trained weight matrix, $\alpha$ is the scaling factor, and $A$, $B$ are the low-rank decomposition matrices.
Through gradient calculation during backpropagation (given loss $L$), we obtain the gradients of parameters $A$ and $B$:
\begin{equation}
    \frac{\partial L}{\partial A} = \alpha \cdot \frac{\partial L}{\partial y} x^\top \cdot B^\top, \quad
    \frac{\partial L}{\partial B} = \alpha \cdot A^\top \cdot \frac{\partial L}{\partial y} x^\top,
\end{equation}
where $\frac{\partial L}{\partial y}$ represents the gradient of the loss with respect to the layer's output $y$.
During optimization, given learning rate $\eta$, the parameter updates follow:
\begin{equation}
    A_{\text{new}} = A - \eta \alpha \cdot \frac{\partial L}{\partial y} x^\top \cdot B^\top, \quad
    B_{\text{new}} = B - \eta \alpha \cdot A^\top \cdot \frac{\partial L}{\partial y} x^\top.
\end{equation}
These update equations reveal a key insight: the magnitude of parameter updates is controlled by the product $\eta \cdot \alpha$.
This indicates that optimizing scaling factors is equivalent to optimizing the learning rate.
For example, doubling the scaling factor $\alpha$ has the same effect as doubling the learning rate $\eta$.

\subsection{NP-hardness of MINLP}
\label{supp_sec_np_hardness}

We show that the MINLP problem defined in {\color{darkred}Section~3.2, Formulation~(6)} is NP-hard through a polynomial-time reduction from the 0-1 knapsack problem, a well-known NP-hard problem.

\noindent\textbf{0-1 Knapsack Problem.} Given a set of $n$ items, each with value $v_i > 0$ and weight $w_i > 0$, and a capacity $W > 0$, find a subset of items maximizing total value while keeping total weight within capacity:
\begin{equation}
    \begin{aligned}
        \max_{\{x_i\}} \quad & \sum_{i=1}^n v_i x_i                                                    \\
        \text{s.t.} \quad    & \sum_{i=1}^n w_i x_i \leq W, \quad x_i \in \{0,1\}, \quad i = 1,\dots,n
    \end{aligned}
\end{equation}

\noindent\textbf{Reduction.} Given any instance of the 0-1 knapsack problem, we construct an instance of the MINLP problem as follows:

\noindent1) Fix the position $s$ to any value $s_0 \in \mathcal{S}$ and scaling factor $\alpha$ to any positive value $\alpha_0$. These values remain constant throughout the reduction.

\noindent2) For each item $i$ in the knapsack problem, create a corresponding layer depth $d_i = i$ where a PEFT module can be inserted. The binary decision $x_i$ of selecting item $i$ maps to whether a PEFT module is inserted at layer $d_i$.

\noindent3) Construct the validation loss function as $\mathcal{L}_{\text{val}} = -\sum_{i=1}^n v_i x_i$.

\noindent4) Map the knapsack constraint to the resource constraint in MINLP as $\sum_{i=1}^n w_i x_i \leq W$, where $W$ represents the computational budget (\eg, maximum number of PEFT modules allowed or FLOPs constraints) in our problem, and then the Formulation~(6) becomes:
\begin{equation}
    \begin{aligned}
        \min_{\{x_i\}} \quad & -\sum_{i=1}^n v_i x_i                                                   \\
        \text{s.t.} \quad    & \sum_{i=1}^n w_i x_i \leq W, \quad x_i \in \{0,1\}, \quad i = 1,\dots,n
    \end{aligned}
\end{equation}

\noindent\textbf{Correctness.} The reduction is polynomial-time as it requires $O(n)$ operations. The solutions between the two problems have a one-to-one correspondence: a solution is feasible in the knapsack problem if and only if it is feasible in the MINLP problem, and the optimal solution to the MINLP problem corresponds to the optimal solution of the knapsack problem (differing only in sign). Since the 0-1 knapsack problem is NP-hard, this reduction proves that Formulation~(6) in the main manuscript is also NP-hard.

\subsection{Discussion: Non-convexity of PEFT Optimization Objective}
\label{supp_sec_conv}
\emph{Theoretically}, our optimization objective is non-convex as it is on deep neural networks (DNN)~\cite{arora2018convergence}. Our optimization of plug-in modules (\textit{e.g.}, LoRA) on DNN is also non-convex.
\emph{Empirically}, extensive prior works have shown that bi-level optimization is effective for non-convex problems of DNN~\cite{liu2021adaptive,liu2023online}, and our results also show consistent and stable model convergence.
In addition, we introduce two techniques to mitigate potential instabilities: 1) a softplus activation to constrain $\gamma$ to be non-negative, and 2) dynamic sampling, which randomly selects 20\% of the training data in meta-training loops to prevent overfitting to local optima.

\section{Method Design and Analysis}
\label{supp_sec_method_design_analysis}

\subsection{Analysis on Data Sampling Strategies in Outer Loop}
\label{supp_sec_outer_loop_sampling}

In {\color{darkred}Section~4.3}, we demonstrate the effectiveness of random sampling in the outer loop of our bi-level optimization framework.
Here, we further explore three alternative sampling strategies for the outer loop: class-balanced sampling, tail-heavy sampling, and larger sampling ratio. Results are provided in Table~\ref{supp_tab_sampling}.

\begin{table}[h]
    \centering
    \small
    \begin{tabular}{l@{\hspace{8pt}}l@{\hspace{4pt}}l@{\hspace{4pt}}l@{\hspace{4pt}}l@{\hspace{4pt}}}
        \toprule[1.3pt]
        Strategy & Head & Med & Tail & Avg \\
        \midrule \midrule
        Random (20\%) & 93.2 & 94.6 & 93.1 & 93.6 \\
        Class-Balanced & 93.0~\textcolor{ForestGreen}{\small$\downarrow$0.2} & 94.5~\textcolor{ForestGreen}{\small$\downarrow$0.1} & 93.3~\textcolor{crimson}{\small$\uparrow$0.2} & 93.6~{\small -} \\
        Tail-Heavy & 92.0~\textcolor{ForestGreen}{\small$\downarrow$1.2} & 93.4~\textcolor{ForestGreen}{\small$\downarrow$1.2} & 94.8~\textcolor{crimson}{\small$\uparrow$1.7} & 93.4~\textcolor{ForestGreen}{\small$\downarrow$0.2} \\
        Random (30\%) & \textbf{93.5}~\textcolor{crimson}{\small$\uparrow$0.3} & \textbf{94.8}~\textcolor{crimson}{\small$\uparrow$0.2} & \textbf{93.9}~\textcolor{crimson}{\small$\uparrow$0.8} & \textbf{94.1}~\textcolor{crimson}{\small$\uparrow$0.5} \\
        \bottomrule[1.3pt]
    \end{tabular}
    \caption{\textbf{Accuracy ($\%$) comparison of different sampling strategies.}
        We compare different sampling strategies in the outer loop of \texttt{MetaPEFT}.
        While tail-heavy sampling achieves the highest tail-class accuracy, it comes at the cost of reduced head-class performance.
        Random sampling with 30\% ratio provides the best overall performance while maintaining balanced improvements across all classes.
        The arrows indicate changes compared to the 20\% random sampling baseline (\textcolor{ForestGreen}{green} for decrease, \textcolor{crimson}{red} for increase).
    }
    \label{supp_tab_sampling}
\end{table}

\noindent\textbf{Class-Balanced Sampling.}
In {\color{darkred}Table~6}, the random sampling strategy preserves the original long-tailed distribution.
Here, we explore a balanced sampling strategy where each class contributes an equal number of samples.
Compared to random sampling, class-balanced sampling achieves comparable overall accuracy (93.8\% vs 93.6\%) but slightly lower head class performance (-0.2\%).

\noindent\textbf{Tail-Heavy Sampling.}
We investigate a tail-heavy sampling strategy where tail classes are sampled with higher probability than head classes (inverse to their original frequencies).
This approach achieves the highest tail-class accuracy (94.8\%) among all sampling strategies, but at the cost of largely reduced head-class performance (-0.6\%).
The trade-off indicates that while emphasizing tail classes during meta-optimization can improve their representation, maintaining some balance with head classes remains important for overall model performance.

\noindent\textbf{Sampling 30\% of Training Data.}
Our observation in {\color{darkred}Table~6} shows that 30\% sampling ratio achieves superior performance, we conduct comparison experiments using this setting.
Results consistently show $0.5\%$ improvement in average accuracy compared to $20\%$ sampling, with particularly strong gains on tail classes ($+0.8\%$).
However, this comes with a $15\%$ increase in meta-optimization time.

\subsection{More Implementation Details}
\label{supp_sec_more_impl_details}

We provide more implementation details as supplementary to {\color{darkred}Section~4.2}.

\subsubsection{Fair Comparisons Between Low-Rank and Adapter-based Methods}
\label{supp_sec_fair_comp}
We compare the architecture difference between LoRA and additive methods, and show how we ensure a fair comparison between them.

\begin{table}[h]
    \centering
    \small
    \begin{tabular}{lccccccc}
        \toprule[1.3pt]
        \multirow{2}{*}{Dataset} & \multirow{2}{*}{Dim} & \multicolumn{2}{c}{LoRA} & \multicolumn{2}{c}{Adapter} & \multicolumn{2}{c}{AdaptFormer} \\
        \cmidrule(lr){3-4} \cmidrule(lr){5-6} \cmidrule(lr){7-8}
        & & Tuner (M) & Head (M) & Tuner (M) & Head (M) & Tuner (M) & Head (M) \\
        \midrule\midrule
        CIFAR100-IR100 & 4 & 0.249 & 0.076 & 0.101 & 0.076 & 0.101 & 0.076 \\
        FUSRS & 4 & 0.197 & 0.005 & 0.136 & 0.005 & 0.136 & 0.005 \\
        DOTA & 4 & 0.249 & 0.011 & 0.101 & 0.011 & 0.101 & 0.011 \\
        Places-LT & 8 & 0.470 & 0.280 & 0.175 & 0.280 & 0.175 & 0.280 \\
        iNat2018 & 256 & 9.437 & 6.253 & 4.749 & 6.253 & 4.749 & 6.253 \\
        \bottomrule[1.3pt]
    \end{tabular}
    \caption{\textbf{Comparison of learnable parameters across different PEFT methods and datasets.}
        ``Dim'' refers to the model dimension (\ie, rank for LoRA, hidden dimension for Adapter/AdaptFormer).
        For each method, ``Tuner (M)'' refers to the number of learnable parameters within the PEFT modules, while ``Head (M)'' refers to those in the classifier head.
    }
    \label{tab_supp_param_count}
\end{table}
\noindent\textbf{Module Size.} We ensure a similar module size between LoRA and adapters. Concretely, we fix the rank of LoRA and the adapter's hidden dimension to a fixed and comparable size.
We follow the setting in~\cite{shi2024lift} to set higher hidden dimensions for larger classifier heads.
The learnable parameters in different methods on different datasets are provided in Table~\ref{tab_supp_param_count}.

\subsubsection{Classifier Head Design and Initialization}
\label{supp_sec_classifier}
We use a linear probing classifier head, which consists of a single linear layer initialized with class-mean features~\cite{snell2017prototypical}.
Specifically, for each class, we compute the mean of its feature representations from the training set (based on the frozen foundation model) and set the corresponding weights of the classifier to these mean vectors.
This initialization leverages the inherent structure of the data, providing a robust starting point that enhances model convergence and performance.

\subsubsection{Evaluation Protocols}
\label{supp_sec_eval_protocol}
We adopt the same evaluation protocols as in~\cite{shi2024lift}.
We report the macro-averaged per-class accuracies for head/medium/tail classes, and report the overall accuracy as the average of these three metrics.
Especially, we use a test-time ensemble (TTE) strategy by averaging the predictions of three independent runs.
While no dedicated literature focuses on TTE, this technique (also named test-time augmentation) has become a de facto standard in many implementations~\cite{wightman2021resnet,shi2024lift}.

\section{Validation Experiments and Observations}
\label{supp_sec_validation_experiments}

\subsection{Generalizing Key Observations to Natural Vision Domain}
\label{supp_sec_lora_pos}

\begin{table*}[h]
    \centering
    \small
    \begin{tabular}{lcccccccccccc}
        \toprule[1.3pt]
        \multirow{2}{*}{Position} & \multicolumn{3}{c}{CIFAR100} & \multicolumn{3}{c}{iNat2018} & \multicolumn{3}{c}{Places-LT} & \multirow{2}{*}{Avg$_\mathrm{tail}$} & \multirow{2}{*}{Avg} \\
        \cmidrule(lr){2-4} \cmidrule(lr){5-7} \cmidrule(lr){8-10}
        & Head & Med & Tail & Head & Med & Tail & Head & Med & Tail & & \\
        \midrule \midrule
        K & 87.3 & 84.6 & 84.0 & 63.2 & 72.3 & 73.3 & 46.5 & 47.0 & 46.4 & 67.9 & 67.9 \\
        Q & 87.4 & 84.3 & 84.1 & 63.3 & 72.7 & 73.7 & 46.7 & 47.2 & 46.6 & 68.1 & 68.1 \\
        V & 91.8 & 88.1 & 85.6 & 65.9 & 74.8 & 75.8 & 47.9 & 47.7 & 45.7 & 69.0 & 70.1 \\
        Out & 91.8 & 88.2 & 85.4 & 65.8 & 75.3 & 75.3 & 47.5 & 47.8 & 46.0 & 68.9 & 70.1 \\
        MLP1 & 92.6 & 87.9 & 85.6 & 66.5 & 75.6 & 76.1 & 48.3 & 47.9 & 45.9 & 69.2 & 70.5 \\
        MLP2 & 92.3 & 88.1 & 86.6 & 65.9 & 75.1 & 76.1 & 47.9 & 47.9 & 45.7 & 69.5 & 70.4 \\
        \bottomrule[1.3pt]
    \end{tabular}
    \caption{\textbf{Accuracy (\%) comparison of intra-block positions in natural vision domain.}
        Results are reported on three transfer scenarios IN21K$\to$\{CIFAR100, iNaturalist-2018, and Places-LT\}.
        ``Out'' denotes the attention output module, and ``MLP1'' and ``MLP2'' denote the first and second FFN linear layers.
        MLP1 achieves the highest overall accuracy (70.5\%), while both MLP positions show strong tail-class performance (69.2-69.5\%).
    }
    \label{tab_position_impact_natural}
\end{table*}
In the main manuscript, our major observations ({\color{darkred}Figure~1c and 1d}) are based on the RS transfer scenario (\ie, IN21K$\to$DOTA).
In this section, we extend our observations in the natural vision domain.

In Table~\ref{tab_position_impact_natural}, we conduct an ablation study on three standard long-tailed benchmarks in the natural vision domain.
We can observe that:
1) MLP positions (MLP1 and MLP2) consistently outperform attention-based positions across all datasets, achieving the highest average accuracy (70.5\% and 70.4\%, respectively).
2) The performance gap is particularly pronounced in tail classes, where MLP positions achieve up to 76.1\% accuracy on iNat2018, higher than attention positions (73.3-75.8\%).
These findings align with our observations in the RS domain, confirming that FFN layers are more effective insertion positions for PEFT modules.

\subsection{Analysis of Position Combination Effects}
\label{supp_sec_position_combinations}
\begin{table*}[h]
    \centering
    \small
    \begin{tabular}{ccccccccccccccc}
        \toprule[1.3pt]
        \multicolumn{4}{c}{Intra-Attn Positions} & \multicolumn{3}{c}{CIFAR100} & \multicolumn{3}{c}{iNat2018} & \multicolumn{3}{c}{Places-LT} & \multirow{2}{*}{Avg$_\mathrm{tail}$} & \multirow{2}{*}{Avg} \\
        \cmidrule(lr){1-4} \cmidrule(lr){5-7} \cmidrule(lr){8-10} \cmidrule(lr){11-13}
        Q & V & K & Out & Head & Med & Tail & Head & Med & Tail & Head & Med & Tail & \\
        \midrule\midrule
        \checkmark & \checkmark &   &   & 91.2 & 87.9 & 86.8 & 66.5 & 75.4 & 76.9 & 48.4 & 47.6 & 45.4 & 69.70 & 69.57 \\
        \checkmark & \checkmark & \checkmark &   & 91.6 & 87.7 & 86.3 & 66.0 & 75.7 & 76.7 & 48.3 & 47.7 & 46.3 & 69.77 & 69.59 \\
        \checkmark & \checkmark & \checkmark & \checkmark & 92.2 & 87.7 & 86.3 & 66.7 & 75.6 & 76.9 & 48.6 & 47.9 & 45.6 & 69.60 & 69.72 \\
        \bottomrule[1.3pt]
    \end{tabular}
    \caption{\textbf{Accuracy (\%) comparison of intra-attention positions' combinations.}
        %
    }
    \label{tab_attention_position_impact}
\end{table*}
\begin{table*}[h]
    \centering
    \small
    \begin{tabular}{cccccccccccccc}
        \toprule[1.3pt]
        \multicolumn{2}{c}{Intra-FFN Positions} & \multicolumn{3}{c}{CIFAR100} & \multicolumn{3}{c}{iNat2018} & \multicolumn{3}{c}{Places-LT} & \multirow{2}{*}{Avg$_\mathrm{tail}$} & \multirow{2}{*}{Avg} \\
        \cmidrule(lr){1-2} \cmidrule(lr){3-5} \cmidrule(lr){6-8} \cmidrule(lr){9-11}
        MLP1 & MLP2 & Head & Med & Tail & Head & Med & Tail & Head & Med & Tail & \\
        \midrule\midrule
        \checkmark &   & 92.6 & 87.9 & 85.6 & 66.5 & 75.6 & 76.1 & 48.3 & 47.9 & 45.9 & 69.20 & 69.60 \\
        & \checkmark & 92.3 & 88.1 & 86.6 & 65.9 & 75.1 & 76.1 & 47.9 & 47.9 & 45.7 & 69.47 & 69.51 \\
        \checkmark & \checkmark & 92.4 & 87.9 & 86.8 & 67.9 & 76.3 & 76.7 & 48.4 & 48.3 & 45.7 & 69.73 & 70.04 \\
        \bottomrule[1.3pt]
    \end{tabular}
    \caption{\textbf{Accuracy (\%) comparison of intra-FFN positions' combinations.}
        %
    }
    \label{tab_mlp_position_impact}
\end{table*}
Beyond the individual position ablations presented in {\color{darkred}Table~2}, we analyze how different positions interact when combined.
We use two types of combinations: 1) combining different positions within attention blocks (Table~\ref{tab_attention_position_impact}), and 2) combining different positions within FFN layers (Table~\ref{tab_mlp_position_impact}).
From the results, we can observe:
1) For attention blocks, adding more positions does not necessarily lead to better performance.
Concretely, using only Q and V achieves comparable or even slightly better tail performance than all positions (69.70\% vs. 69.60\% in Table~\ref{tab_attention_position_impact} rows 1 and 3).
2) For FFN layers, combining both MLP1 and MLP2 positions shows clear advantages in Table~\ref{tab_mlp_position_impact}.
The combined approach achieves the best overall performance (70.04\%) and tail performance (69.73\%), outperforming single position variants (69.60\% and 69.51\% for MLP1 and MLP2, respectively).
3) The improvements from position combinations are more pronounced in FFN layers compared to attention blocks.
Concretely, FFN combinations show a $+0.53\%$ improvement in average accuracy (from 69.51\% to 70.04\% in Table~\ref{tab_mlp_position_impact}), while attention combinations only yield $+0.15\%$ improvement (from 69.57\% to 69.72\% in Table~\ref{tab_attention_position_impact}).

\begin{table*}[h]
    \centering
    \small
    \begin{tabular}{lccccccccccc|c}
        \toprule[1.3pt]
        \multirow{2}{*}{Method} & \multicolumn{3}{c}{CIFAR100} & \multicolumn{3}{c}{iNat2018} & \multicolumn{3}{c}{Places-LT} & \multirow{2}{*}{Avg} \\
        \cmidrule(lr){2-4} \cmidrule(lr){5-7} \cmidrule(lr){8-10}
        & Head & Med & Tail & Head & Med & Tail & Head & Med & Tail & \\
        \midrule \midrule
        LoRA & 91.2 & 87.9 & 86.8 & 55.8 & 64.0 & 64.7 & 47.5 & 47.8 & 46.1 & 66.5 \\
        AdaptFormer & 92.3 & 88.3 & 86.8 & 68.7 & 76.7 & 78.0 & 49.1 & 47.9 & 45.1 & 71.1 \\
        LoRA+AdaptFormer & 92.5 & 88.4 & 87.0 & 67.9 & 77.2 & 77.8 & 48.6 & 48.0 & 45.3 & 71.2 \\
        \bottomrule[1.3pt]
    \end{tabular}
    \caption{\textbf{Accuracy (\%) of combining different additive PEFT methods.}
        %
    }
    \label{tab_compare_ensemble}
\end{table*}

We also compare the ensemble of different PEFT methods in Table~\ref{tab_compare_ensemble}.
Results show that simply combining LoRA and AdaptFormer yields marginal improvements ($+0.1\%$ on average) over using AdaptFormer alone, and shows little gain for tail classes (\eg, 77.8\% vs 78.0\% on iNat2018).
This suggests that a naive or straightforward combination of PEFT methods is ineffective.

\subsection{Hyperparameter Optimization is Particularly Crucial for Tail Classes}
\label{supp_grid_search}
\begin{figure}[h]
    \centering
    \includegraphics[width=0.98\textwidth]{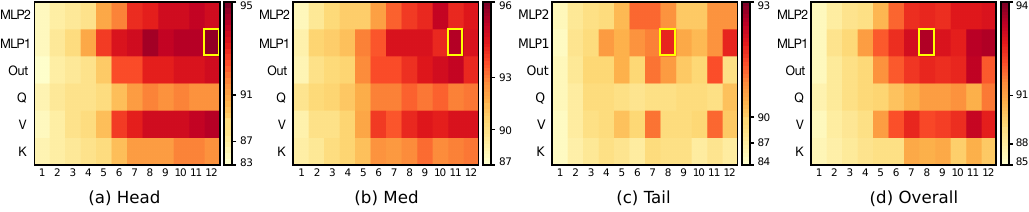}
    \caption{\textbf{Accuracy heatmap for head/med/tail/overall classes.}
        Visualization of accuracy heatmaps on different intra-block layers vs among different blocks (depth), where the block stands for the attention block of ViT.
        Results are reported on transfer scenarios IN21K$\to$DOTA.
        The optima in each heatmap is highlighted by a yellow box.
        Results show that:
        1) the (c) Tail classes exhibit non-monotonic accuracy changes across positions, while (a) Head and (b) Medium classes show monotonic trends;
        2) the optimal configuration in (c) determines the overall optimal configuration in (d), indicating that tail-class performance dominates the model's overall performance.
        These findings validate our position-aware optimization strategy for long-tailed datasets.
    }
    \label{supp_fig_heatmap_all_pos}
\end{figure}
The heatmaps in {\color{darkred}Figure~1} present the average accuracy across all classes.
In this section, we provide a detailed breakdown of head/medium/tail/overall class performance for each grid search configuration in Figure~\ref{supp_fig_heatmap_all_pos}.

Our analysis reveals that hyperparameter optimization is particularly challenging yet crucial for tail classes.
First, for head classes (Figure~\ref{supp_fig_heatmap_all_pos}(a)) and medium classes (Figure~\ref{supp_fig_heatmap_all_pos}(b)), the accuracy shows a monotonic increase pattern towards deeper layers (8-12) and upper positions (MLP1/MLP2).
In contrast, for tail classes, the accuracy values do not show a clear correlation with either depth or position (Figure~\ref{supp_fig_heatmap_all_pos}(c)).
Second, the performance on tail classes determines the model's overall performance: the optimal configuration (MLP1 position and depth 8) in Figure~\ref{supp_fig_heatmap_all_pos}(c) directly corresponds to the overall optimal configuration in Figure~\ref{supp_fig_heatmap_all_pos}(d).
These findings highlight the importance of our optimization method for long-tailed datasets.

\subsection{Analysis of LoRA Rank: Rank Optimization Can be Decoupled from Other Hyperparameters}
\label{supp_sec_lora_rank}

\begin{figure}[h]
    \centering
    \includegraphics[width=0.60\textwidth]{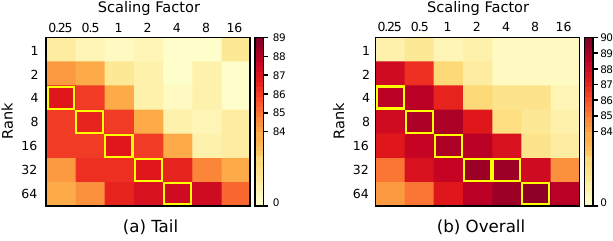}
    \caption{\textbf{Linear relationship between rank and optimal scaling factor.}
        The heatmaps show accuracy distributions across different ranks and scaling factors for (a) tail classes and (b) all classes.
        Results reported on IN21K$\to$CIFAR100.
        Yellow highlights indicate the optimal scaling factor for each rank (the rank 32 in Figure~(b) has two equally optima).
        We can observe a consistent linear relationship, \ie, optimal scaling factor $\propto$ rank.
        This pattern suggests that rank optimization can be decoupled from remaining hyperparameters.
    }
    \label{supp_fig_heatmap_rank}
\end{figure}
We used fixed ranks in experiments ({\color{darkred}Section~4.3}) because our method generalizes across different rank settings.
We validate this claim through two aspects:
1) rank's impact on model performance, and
2) consistency of position/scaling factor observations across different ranks (Figure~\ref{supp_fig_heatmap_rank}).

Concretely, from Figure~\ref{supp_fig_heatmap_rank}, we have two key findings:
1) Rank selection significantly impacts model performance. For instance, in Figure~\ref{supp_fig_heatmap_rank}(b), at scaling factor 4, changing the rank from 1 to 32 improves accuracy from 1.0\% to 89.2\% (dramatically +88.2\%).
2) Rank and optimal scaling factor has a linear relationship.
Specifically, the optimal scaling factor can be directly computed as $\frac{\text{rank}}{n}$, where $n$ is a dataset-specific constant (\eg, $n$=16 for CIFAR100-IR100 in Figure~\ref{supp_fig_heatmap_rank}).
This linear relationship suggests that the optimization of rank and remaining hyperparameters can be decoupled.
Given these observations, our \texttt{MetaPEFT} method optimizes position and scaling factor hyperparameters while keeping the rank fixed.

\subsection{t-SNE Feature Visualization}
\label{supp_sec_tsne}
We provide two sample figures: t-SNE maps for (a) LoRA and (b) LoRA + Ours. LoRA + Ours achieves better class separation and more compact clusters. We merge head categories and tail categories for clearer visualization, and blue/red circles are head/tail classes, respectively. From the figure, we could observe better class separation and more compact clusters in each class. Complementary to feature-level visualization, concept-based interpretation methods~\cite{yu2025coe} could provide additional insights into how PEFT modules affect model decision-making.

\begin{figure}[h]
    \centering
    \includegraphics{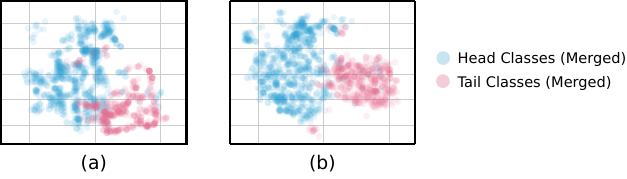}
    \caption{\textbf{t-SNE Visualization} of (a) LoRA and (b) LoRA + Ours.}
    \label{fig:enter-label}
\end{figure}

\subsection{Apply to More LoRA/Ensembled Methods}
\label{supp_sec_ensemble}
The Table~6 of LIFT~\cite{shi2024lift} shows that AdaptFormer and LoRA outperform other PEFT methods in long-tailed data distributions, especially for tail classes, so we adopt AdaptFormer/Adapter/LoRA as our baseline methods. Table~4 shows we beat LoRA by 1.2\% for tail classes and 1.13\% on average.
Additionally, in Table~\ref{tab:liftlotr}, we apply our meta-optimization strategy to LIFT and recent low-rank methods LoTR~\cite{bershatsky2024lotr} on IN21K$\to$DOTA.
Our method shows limited improvement on LIFT, which is expected as LIFT is already an excellent work with well-tuned scaling factors in its design.

\begin{table}[h]
    \centering
    \small
    \begin{tabular}{lccc}
        \toprule[1.3pt]
        \multirow{2}{*}{Method} & \multicolumn{3}{c}{IN21K $\to$ DOTA} \\
        \cmidrule(lr){2-4}
        & Head & Med & Tail \\
        \midrule \midrule
        LIFT~\cite{shi2024lift} & 93.1 & 94.9 & 90.2 \\
        \cellcolor{gray!20}\quad w/ Ours & \cellcolor{gray!20}93.2 & \cellcolor{gray!20}94.7 & \cellcolor{gray!20}90.6 \\
        LoTR~\cite{bershatsky2024lotr} & 93.0 & 93.1 & 90.3 \\
        \cellcolor{gray!20}\quad w/ Ours & \cellcolor{gray!20}93.1 & \cellcolor{gray!20}92.9 & \cellcolor{gray!20}91.1 \\
        \bottomrule[1.3pt]
    \end{tabular}
    \caption{\textbf{Apply our method to LIFT and LoTR.}}
    \label{tab:liftlotr}
\end{table}

\section{Limitations and Future Works}
\label{supp_sec_limits}
Despite the promising results of \texttt{MetaPEFT}, future works remain. First, its scalability to larger models and more diverse RS spectrum datasets warrants further investigation. Currently, evaluation is limited to the FUSRS SAR dataset due to the scarcity of SAR data, highlighting the need for larger, standardized SAR recognition benchmarks.
Second, the current framework relies on a fixed backbone architecture (\textit{e.g.}, ViT-B/16). Exploring how \texttt{MetaPEFT} generalizes across different backbone architectures, such as CNN models or hybrid transformer-CNN models, can broaden its applicability.
Third, our bi-level optimization framework reduces overfitting in tail classes, however, it introduces additional computational overhead in outer loops.
Fourth, our bi-level optimization framework still depends on several outer loop hyperparameters, including meta-parameter learning rate, learning steps, early stop, and update frequency. Reducing them remains an important direction for future research.
Our future work could investigate lightweight optimization techniques or meta-learning strategies that reduce computational costs without compromising model performance.

As part of our ongoing work, some more interesting explorations (\textit{e.g.}, meta optimization under noisy data and insufficient data challenges, or integrating with advanced reasoning strategies for vision models~\cite{zhao2025unsupervised}) are currently ongoing. We will share our progress on our GitHub repository.

\end{document}